\documentclass{article} 
\usepackage{iclr2019_conference,times}

\usepackage{hyperref}
\usepackage{url}

\usepackage{textcomp}
\usepackage{graphicx}
\usepackage{multirow}
\usepackage{minibox}
\usepackage{xcolor,colortbl}
\usepackage{balance}
\usepackage{color}
\usepackage{amsmath}
\usepackage{mathtools}

\usepackage[numbers]{natbib}

\usepackage[flushleft]{threeparttable}

\usepackage{bm}
\usepackage{amsfonts}

\DeclareMathOperator*{\argmax}{max}
\DeclareMathOperator*{\argmin}{min}

\usepackage[flushleft]{threeparttable}

\definecolor{Gray}{gray}{0.95}
\definecolor{White}{gray}{1.0}
\newcolumntype{g}{>{\columncolor{Gray}}l}

\title{Applying Adversarial Auto-encoder for Estimating Human Walking Gait Quality Index}

\author{Trong-Nguyen Nguyen \& Jean Meunier\\
DIRO, University of Montreal\\
\texttt{\{nguyetn, meunier\}@iro.umontreal.ca}
}

\iclrfinalcopy 
\begin{document}

\maketitle

\begin{abstract}
This paper proposes an approach that estimates human walking gait quality index using an adversarial auto-encoder (AAE), i.e. a combination of auto-encoder and generative adversarial network (GAN). Since most GAN-based models have been employed as data generators, our work introduces another perspective of their application. This method directly works on a sequence of 3D point clouds representing the walking postures of a subject. By fitting a cylinder onto each point cloud and feeding obtained histograms to an appropriate AAE, our system is able to provide different measures that may be used as gait quality indices. The combinations of such quantities are also investigated to obtain improved indicators. The ability of our method is demonstrated by experimenting on a large dataset of nearly 100 thousands point clouds and the results outperform related approaches that employ different input data types.
\end{abstract}

\section{Introduction}
Gait analysis has a wide variety of applications in medicine, person identification or activity recognition. In healthcare, many gait measurements can be done for the precise identification of locomotion problems and the planning of an appropriate treatment. However there are many situations where an overall measurement of the quality of gait would be useful to the clinician. In this work, we propose such gait quality index using a computer vision approach and adversarial auto-encoder.

\subsection{Common computer vision approaches for gait analysis}

In order to deal with problems of gait analysis with computer vision methods, researchers employed different data types. Early studies started with a color camera that captures subject silhouettes under a specific view point. Many gait signatures have been introduced to describe various properties of each individual. For example, the Motion History Image (MHI)~\cite{Davis2001} used the pixel intensity to represent the motion history at the corresponding location. Another gait signature, Gait Energy Image (GEI)~\cite{Han2006}, focused on person identification by calculating an average image of consecutive aligned silhouettes. Beside such characteristics, researchers also proposed some problem-oriented features describing the movement. By proposing a 4-d vector that employed the MHI to indicate subject posture in each frame,~\citet{Nguyen2014extracting} measured the walking gait index for each gait cycle as the log-likelihood provided by a hidden Markov model (HMM). Differently from that work,~\citet{Bauckhage2009} captured the walking silhouettes under the frontal view in order to detect abnormal gaits via the balance deficiency of motion. A common drawback of such silhouette-based gait analysis is the significant dependency on the camera view point and self-occlusion in captured silhouettes.

Another popular input of gait analysis systems is 3D skeleton. Since the Kinect 1 and 2 were released by Microsoft with low prices and SDK for skeleton localization~\cite{Shotton2011realtime,Shotton2013efficient}, these devices have been applied in many studies surpassing previous approaches using a 2D skeleton or other 2D model. Such skeletons have been demonstrated to be useful for a wide variety of applications such as recognizing predefined gaits~\cite{Jiang2015}, analyzing pathological gaits~\cite{Bigy2015}, and detecting abnormal gaits~\cite{Nguyen2016}. However, these skeletons that are detected based on depth maps may have a higher risk of posture deformation with pathological gait e.g. due to self-occluded parts.

\begin{figure}[t]
\centering
\scalebox{1}{
\begin{picture}(404,140)
	\put(0,14){\includegraphics[width=0.4\textwidth]{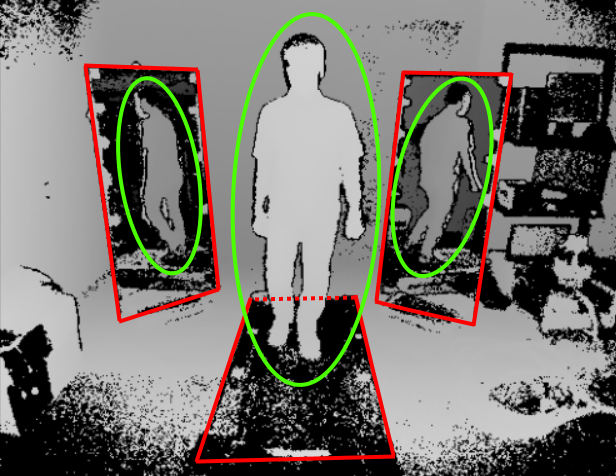}}
	\put(184,14){\includegraphics[width=0.525\textwidth]{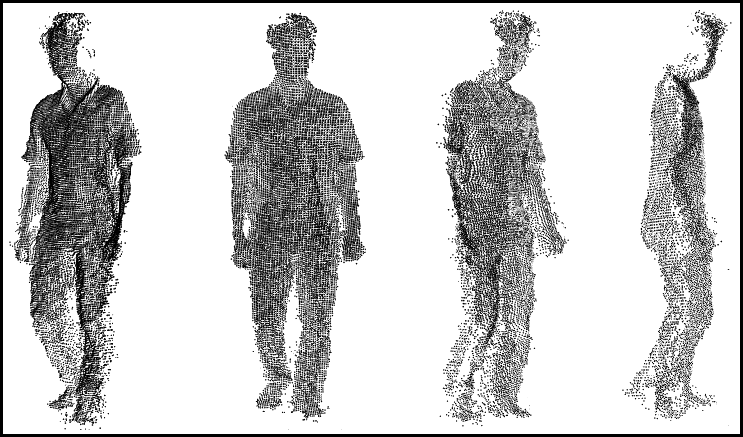}}
	\put(4,0){(a) Depth map captured by our system}
	\put(220,0){(b) Reconstructed point cloud}
\end{picture}}
\caption{Data acquisition of our system: (a) a depth map showing our setup that includes a treadmill and two mirrors (highlighted by rectangles), each depth map captures three subject's surfaces (marked by ellipses) under different view points, (b) a reconstructed point cloud of a similar posture.}
\label{fig:setup}
\end{figure}

To alleviate the previous problems, our method attempts to represent a subject pose by 3D information collected from different view points. The effect of view point dependency (including self-occlusion) would thus be reduced. Instead of employing a system of multiple cameras as in~\cite{Auvinet2012,Lopez2016}, we use only one Time-of-Flight (ToF) depth camera together with two mirrors. Each mirror plays the role of a virtual depth camera where its position is symmetric with the real one through the corresponding mirror plane. A depth map captured by the ToF camera in our setup is presented in Fig.~\ref{fig:setup}. Since the scene is captured by only one device, the task of camera synchronization is thus avoided. Furthermore, the system is not expensive and does not require precise placement of sensors or markers on the body of the patient (e.g. motion capture). Our system provides a 3D point cloud of a subject walking on a treadmill for each depth frame using the method proposed in~\cite{NguyenReportKinect2,NguyenKinect2}. These point clouds are then fed to the AAE (next section) to obtain a gait quality index.

\subsection{Adversarial auto-encoder}\label{sec:AAE}

\begin{figure}[t]
\centering
\begin{picture}(250,120)
	\put(0,10){\includegraphics[scale=0.667]{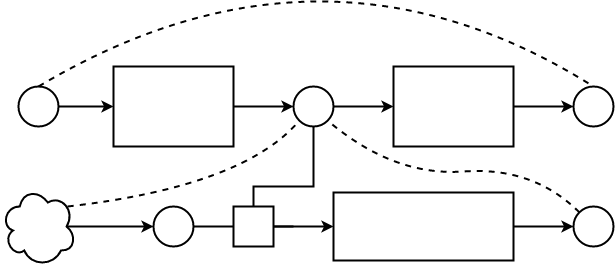}}
	\put(10.5,71.5){$X$}\put(234,70.6){$\widehat{X}$}\put(123.5,73){$z$}
	\put(54,73){\minibox{encoder\\\hspace{0.45cm}\textit{Q}}}\put(166,73){\minibox{decoder\\\hspace{0.45cm}\textit{P}}}
	\put(12.5,22){\textbf{P}}\put(67,23.5){$\tilde{z}$}
	\put(98,23.5){$\cup$}\put(144,23.5){\minibox{discriminator\\\hspace{0.75cm}\textit{D}}}
	\put(236,23.5){$p$}
	\put(110,44.5){$l^-$}\put(81,28.5){$l^+$}
\end{picture}
\caption{A typical AAE where $X$ and $\widehat{X}$ are respectively an input and its reconstruction result provided by the AE, $z$ is the representation of $X$ in latent space, \textbf{P} is a predefined prior distribution that draws samples $\tilde{z}$, $l^+$ and $l^-$ respectively indicate the assigning of positive and negative labels, and $p$ is the probability that an input is real, i.e. its label is positive ($l^+$). The operation $\cup$ represents the union of labeled samples $z$ and $\tilde{z}$. In this diagram, the dash lines indicate components that may provide partial measures.}
\label{fig:overview}
\end{figure}
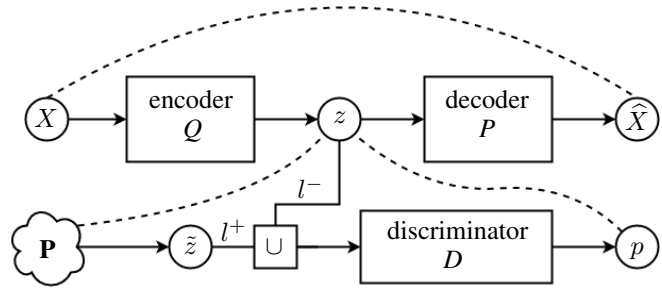

An AAE can be considered as a combination of an auto-encoder (AE) and a generative adversarial network (GAN)~\cite{Goodfellow2014}. The AAE was introduced in~\cite{Alireza2016} to perform variational inference so that the aggregated posterior of latent variables is similar to a given prior distribution. That model focuses on supporting the task of sample generation that is currently a research trend. Our work, however, considers the AAE under another perspective. Inspired by recent works~\cite{Yu2010,Nguyen2018BHI} where a weighted combination of partial measures helped to improve the final assessment, we believe that an AAE could be applied in the same fashion since it contains multiple partial networks that can provide input-oriented measures. Our system does not focus on evaluating generated samples, the objective instead is to tune model weights so that such partial measures are reasonable to indicate the gait index for each input of point cloud. An overview of the AAE used in this work is presented in Fig.~\ref{fig:overview}.

The remainder of this paper is organized as follows: Section~\ref{sec:method} describes the processing flow of our approach; the experiments on a large dataset and a comparison with related methods are given in Section~\ref{sec:experiment}; Section~\ref{sec:conclusion} presents the conclusion together with possible extensions that may improve the current work.

\section{Proposed method}\label{sec:method}

As presented in Fig.~\ref{fig:overview}, the input $X$ is fed to an AE where the number of units in the input layer is fixed, the point clouds should thus be converted into an appropriate representation. In other words, such point clouds need to be normalized to vectors or images (depending on the AE structure) with a predefined length or resolution. Differently from studies~\cite{Bauckhage2009,Nguyen2016} where the temporal factor was directly integrated into the stage of feature extraction, we first perform the gait index measurement on each individual point cloud and then consider a sequence of such measures to assess the whole gait.

\subsection{Posture representation}\label{sec:hist}
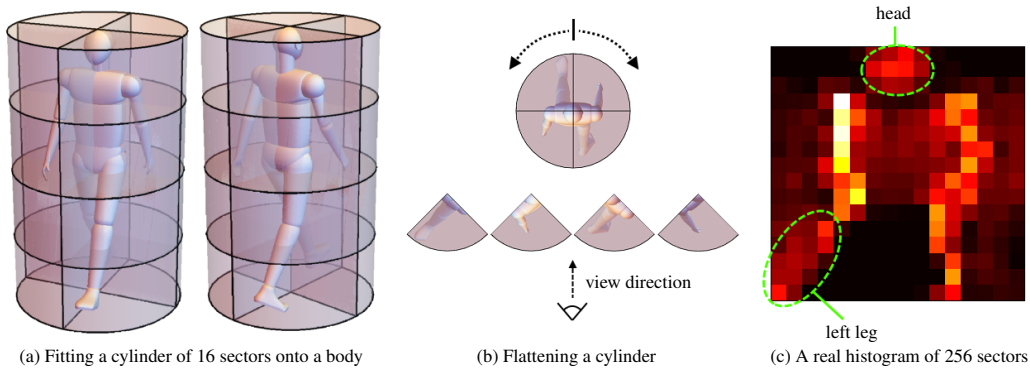
\begin{figure*}[t]
\centering
\scriptsize
\scalebox{1}{
\begin{picture}(404,150)
	\put(0,10){\includegraphics[width=\textwidth]{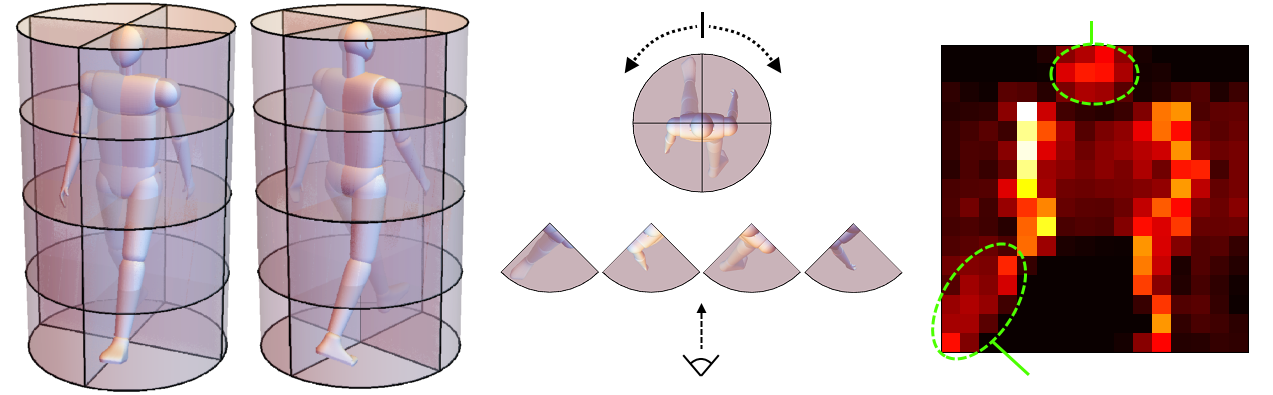}}
	\put(10,0){(a) Fitting a cylinder of 16 sectors onto a body}
	\put(183,0){(b) Flattening a cylinder}
	\put(294,0){(c) A real histogram of 256 sectors}
	\put(334,130){head}\put(315,9){left leg}\put(224,28){view direction}
\end{picture}}
\caption{Illustration of estimating a cylindrical histogram: (a) a cylinder, that contains 16 equal-volume sectors, is employed to segment a 3D point cloud (a 3D model was used in the figure to provide a better visualization), (b) the collection of cylindrical sectors is then flattened to give a 2D representation, i.e. a histogram where each bin is the number of 3D points inside the corresponding sector, and (c) a pseudo-color version of such histogram (of size $16 \times 16$) that was estimated from our real data. Human model created by Dano Vinson (\url{https://grabcad.com}).}
\label{fig:hist}
\end{figure*}

Each input of our AAE is a 3D point cloud that is reconstructed from the corresponding depth map using the method~\cite{NguyenKinect2} [Fig.~\ref{fig:setup}(b)]. In order to normalize the input representation, we use a cylinder with same-size 3D sectors to fit the point cloud. The main axis of the cylinder goes through the cloud centroid and is normal to the ground plane (or treadmill surface in our experiments). The top and bottom bases respectively go through the highest and lowest points (along the main axis) of the cloud. The cylinder's radius is large enough to guarantee that every point is inside the cylinder volume. The collection of such 3D sectors can be flattened to obtain a 2D histogram where each bin value indicates the number of 3D points belonging to the corresponding sector.

An illustration of our histogram formation is shown in Fig.~\ref{fig:hist}. First, a cylinder is employed to fit the input 3D point cloud according to the mentioned constraints (i.e. the main axis, top and bottom bases, and the radius). The cylinder is then separated into same-size sectors using horizontal and vertical slices as shown in Fig.~\ref{fig:hist}(a). It is obvious that the cylinder's main axis is normal to the horizontal slices and is the intersection of the vertical ones. In the next step, the number of 3D points inside each sector is counted, the input point cloud thus becomes a cylindrical histogram. In order to get an appropriate representation, the collection of sectors is flattened to a typical 2D array. The flattening also provides a visual understanding since body parts can be easily localized on the histogram (see Fig.~\ref{fig:hist}(c) where the head and the left leg are indicated). Let us notice that in our work and experiments, this histogram is seen from the back as shown in Fig.~\ref{fig:hist}(b). Such arrangement of sectors is not strictly a constraint because our model does not consider this factor. In the implementation stage, such cylindrical histogram can be formed by performing a loop on 3D points and determining the corresponding sectors based on geometric calculations.

After estimating the histogram, an enhancement is performed according to the following reasons. First, the value assigned to each bin is the number of points belonging to the corresponding sector, measuring gait index directly on such data is thus significantly affected by the subject's shape properties. For example, the cloud that is formed with a fat subject should contain much more points than a thin one. Therefore, a normalization is necessary. Each histogram is thus scaled to the range [0, 1]. This operation is also useful for further processing where neural networks are employed. Beside the scaling, the output range is also separated into 256 levels. We empirically found that this step slightly improves the accuracy of our model for the task of gait index measurement. The histogram can thus be stored and directly visualized as a typical image. In our work, the selected size of cylindrical histogram is $16 \times 16$. Notice that this is just an arbitrary choice, not necessarily the optimal one.

\subsection{Model components}\label{sec:model}

In this work, the AAE is our choice for building the model because we focus on unsupervised learning. Since there are numerous possible walking gaits, collecting patterns of every type of gait for a supervised learning is nearly impossible. On the other hand, the unsupervised learning does not consider the data label and is appropriate for a training set that contains samples belonging to only one class. Our idea is to create a model that provides the score measuring the similarity between an input and known gaits. Another reason for the choice of unsupervised learning is that gait indices are usually used to assess the normality of a subject walking, a one-class classifier is thus appropriate. In our experiments, the AAE was trained using only normal walking gaits.

As visualized in Fig.~\ref{fig:overview}, our model contains 3 main partial networks: the encoder and decoder that belong to the AE, and the discriminator that estimates the probability that an input is drawn from the given distribution $\mathbf{P}$. Each network is simply designed as a stack of fully-connected layers. Unlike popular deep learning models, we do not use any convolutional layer in our AAE because of the following reason. The input $X$ is a normalized histogram instead of a natural image. Different inputs have a similar structure (e.g. body part position, body orientation), a convolutional layer (as well as a pooling layer) is thus not necessary to highlight common low-level features. In our work, each input sample $X$ contains 256 elements (corresponding to a histogram of size $16 \times 16$), and the latent space (i.e. $z$) has 16 dimensions. The structures of the three partial networks are presented in Table~\ref{table:structure}.
\begin{table*}[t]
\centering
\caption{Structures of the 3 partial networks in our AAE.}
\label{table:structure}
\begin{threeparttable}
\begin{tabular}{cc|cc|cc}
\hline
\multicolumn{2}{c|}{\textbf{encoder} $\bm{Q(z|X)}$} & \multicolumn{2}{c|}{\textbf{decoder} $\bm{P(\widehat{X}|z)}$} & \multicolumn{2}{c}{\textbf{discriminator} $\bm{D(z)}$} \\ 
layer & no. of units & layer & no. of units & layer & no. of units \\ \hline \hline
input & 256 & input & 16 & input & 16 \\ \hline
fc & 96 & fc & 96 & fc & 96 \\ 
lrelu & - & lrelu & - & lrelu & - \\ \hline
fc & 16 & fc & 256 & fc & 1 \\ 
 & & sigmoid & - & sigmoid & - \\ \hline
\end{tabular}
\begin{tablenotes}
  \item Abbreviation: fc = fully-connected, lrelu = leaky ReLU
\end{tablenotes}
\end{threeparttable}
\end{table*}

The three components in our AAE use a similar hidden layer of (experimentally selected) 96 units that are fully connected from the input and are then activated by a leaky ReLU (rectified linear unit). The output layer of the decoder $P$ attempts to reconstruct the input $X$ of the AE. Therefore, 256 units are contained in that layer and followed by the sigmoid activation to guarantee each outputted element asymptotically belongs to the range [0, 1]. The sigmoid in the discriminator $D$ focuses on another objective that is to estimate a probability.

Our training stage employed three different optimizers. The first one uses the Adam algorithm~\cite{Kingma2014} to train the encoder $Q$ and decoder $P$ together as a typical AE to minimize the reconstruction error. The loss function is cross entropy as follows:
\begin{equation}
	L_{AE} = -X\mathrm{log}(\widehat{X}) - (1-X)\mathrm{log}(1-\widehat{X})
	\label{eq:loss_AE}
\end{equation}
where the input terms are similar to the notations in Fig.~\ref{fig:overview}. The two remaining optimizers deal with two components of the adversarial loss that has the overall form:
\begin{equation}
	\argmin_Q\argmax_D {\mathop{{}\mathbb{E}}}_{\tilde{z}\sim\mathbf{P}}[\mathrm{log}D(\tilde{z})] + {\mathop{{}\mathbb{E}}}_{z\sim Q(z|X)}[\mathrm{log}(1-D(Q(z|X)))]
	\label{eq:loss_GAN}
\end{equation}
where \textbf{P} is the given prior distribution and the encoder $Q(z|X)$ plays the role of the generator in the GAN. The optimization of such minimax function can be performed by alternatively optimizing the two following losses:
\begin{equation}
	\begin{split}
	L_D = \frac{1}{2n}\sum_{i=1}^{n}[-\mathrm{log}D(\tilde{z}_i)-\mathrm{log}(1-D(Q(z_i|X_i)))] \\+ \frac{\gamma}{2}R_D(\tilde{z},z,D)
	\end{split} 
	\label{eq:loss_D}
\end{equation}
\begin{equation}
	L_Q = \frac{1}{n}\sum_{i=1}^{n}[-\mathrm{log}D(Q(z_i|X_i))]
	\label{eq:loss_Q}
\end{equation}
where $n$ is the number of samples $\tilde{z}$ with positive label drawn from \textbf{P} as well as the number of normal gait postures $X$ drawn from the training set. $\gamma$ is an annealing factor that is combined with the regularization $R_D$ in order to increase the stability when training the discriminator~\cite{Roth2017}. The two losses $L_D$ and $L_Q$ were respectively optimized using SGD and Adam algorithms in our experiments. Both losses are opposing functions, $L_D$ updates the discriminator to better differentiate positive samples $\tilde{z}$ generated by \textbf{P} from negative samples $z$ computed by the encoder while $L_Q$ updates the GAN generator, which is also the encoder of the AE, to fool the discriminator.

\subsection{Gait index estimation}\label{sec:measure}
As mentioned in Section~\ref{sec:AAE}, our gait index is estimated as a combination of measures obtained from partial networks. The first measure is the reconstruction loss $\Upsilon_{AE}$ that is estimated as the Root-Mean-Square Error (RMSE) between an input $X$ and its output $\widehat{X}$. The second operand of the combination is the probability $\Upsilon_\mathbf{P}$ that $z$ is sampled from the prior distribution \textbf{P}. This is a reasonable consideration since we expect that the AAE forces the distribution of trained latent variables $z$ being similar to \textbf{P}, a mapped $Q(z|X)$ of an abnormal gait posture should thus belong to a region of low probability density. It is noticeable that a range normalization is necessary for $\Upsilon_\mathbf{P}$ to obtain a measure belonging to [0, 1]. The last measure, notated as $\Upsilon_D$, is the output $p=D(z)$ of the discriminator. Concretely, the discriminator $D$ should assign high values to normal walking postures and lower values to ones that are different from training samples since $D$ has been fooled to consider the latent representation $z$ of a normal posture as a positive sample.

It is obvious that the three terms $\Upsilon_{AE}$, $\Upsilon_\mathbf{P}$ and $\Upsilon_D$ are non-negative, but the posture orders corresponding to these values are not the same. For example, a (very) normal posture should provide $\Upsilon_{AE}$ that tends to be near the low-end, while $\Upsilon_\mathbf{P}$ and $\Upsilon_D$ should be near the high-end of their range. The combination of the three measures is calculated according to a weighted sum as
\begin{equation}
	\begin{split}
	\Upsilon_X &= w_{AE}\Upsilon_{AE} + w_\mathbf{P}\Upsilon_\mathbf{P} + w_D\Upsilon_D\\
						&=w_{AE}\frac{\|X-\widehat{X}\|_2}{\sqrt{m_X}} + w_\mathbf{P}f_s(Q(z|X)|\mathbf{P}) + w_DD(Q(z|X))
	\end{split}
	\label{eq:combination}
\end{equation}
where $m_X$ is the dimension of $X$ and $f_s$ is a range scaling operation that applies on a probability density function $f$ as $f_s(Q(z|X)|\mathbf{P}) = \frac{f(Q(z|X)|\mathbf{P})}{f(0|\mathbf{P})}$. The denominator scales the output of $f$ to the range [0, 1]. In our experiments, $m_X$ was 256 since the size of cylindrical histograms was $16 \times 16$, and the prior distribution \textbf{P} was a multivariate normal distribution with zero mean and scalar covariance matrix. Therefore, $f(0|\mathbf{P})$ corresponds to the maximum value of $f$.

An unknown factor in eq.~(\ref{eq:combination}) is the weight values. We consider the combination of 2 and 3 quantities. The removal of a measure in the former case is performed by simply assigning its weight to zero in eq.~(\ref{eq:combination}). 
Since the three terms $\Upsilon_{AE}$, $\Upsilon_\mathbf{P}$ and $\Upsilon_D$ are normalized in the range [0, 1], the weight of $\Upsilon_i$ is computed as $w_i = \frac{\sum_i{s_i}}{s_i}$ where $s_i$ is the average value of the corresponding measure $m_i$ calculated from training patterns as in~\cite{Nguyen2018BHI}. In other words, the weight calculation of a measure only depends on its values obtained in the training stage. The numerator is a constant in all the weights to facilitate the computation. 
After obtaining the weights, the gait index of a posture (i.e. a cylindrical histogram) is calculated according to eq.~(\ref{eq:combination}). The combination is expected to improve the gait quality measure as follows. In the three measures $\Upsilon_{AE}$, $\Upsilon_\mathbf{P}$ and $\Upsilon_D$, the first one is the most significant factor since many studies demonstrated the ability of auto-encoder in anomaly detection (e.g.~\cite{Martinelli2004,Sakurada2014}). This property is embedded into eq.~(\ref{eq:combination}) by $w_{AE}$ that is much greater than $w_\mathbf{P}$ and $w_D$. Therefore, $\Upsilon_\mathbf{P}$ and $\Upsilon_D$ should be considered as additional factors to enhance the main indicator $\Upsilon_{AE}$.

\section{Experiments}\label{sec:experiment}

\subsection{Dataset}\label{sec:dataset}
In order to evaluate the proposed method, we performed the gait index estimation [eq.~(\ref{eq:combination})] on a dataset of 9 types of walking gaits including normal and abnormal ones that reduce the gait balance. Concretely, a sole with 3 different thicknesses (5/10/15 \textit{cm}) was padded under one of the two feet to simulate frontal asymmetry. The two remaining gait types were performed by attaching a weight of 4 \textit{kg} to an ankle to impair the walking speed on one side of the body. The dataset was acquired by 9 subjects and the camera frame rate was 13 fps. Each gait of a subject was captured as a sequence of 1200 point clouds, 1200 frontal silhouettes and 1200 skeletons, synchronously. Details of the dataset can be found in~\cite{NguyenReportDataset}\footnote{~This dataset is available online at \url{http://www.iro.umontreal.ca/~labimage/GaitDataset}}.

\subsection{Assessment scheme}\label{sec:assessment}
The evaluation was performed by considering gait indices in the task of distinguishing normal and abnormal gaits. The dataset was split into training and test sets under two schemes. The first one used the default separation suggested in~\cite{NguyenReportDataset} where the gaits of 5 subjects are available for the training stage, and the test set contains the 4 remaining ones. The other evaluation scheme is to perform leave-one-out (on subject) cross-validation to get a more general assessment. We also reimplemented related works (including~\cite{Bauckhage2009,Nguyen2016,Nguyen2018BHI}) that employ different data types to provide a comparison. These studies used various quantities for evaluation: classification accuracy in~\cite{Bauckhage2009}, Area Under Curve (AUC) of the Receiver Operating Characteristic (ROC) curve in~\cite{Nguyen2016} and Equal Error Rate (EER) in~\cite{Nguyen2018BHI}. We used the EER to indicate the ability of each method since this is related to the classification error and is estimated according to the ROC curve. Beside the per-frame assessment, the temporal factor was also considered by using the average measure over (non-overlapping) segments of consecutive frames as the gait indices. Such segment-based measure is usually considered as a better gait index indicator compared with the per-frame one as reported in~\cite{Bauckhage2009,Nguyen2016,Nguyen2018BHI}.

\begin{figure}[t]
\centering
\begin{picture}(250,154)
	\put(-2,0){\includegraphics[scale=0.59]{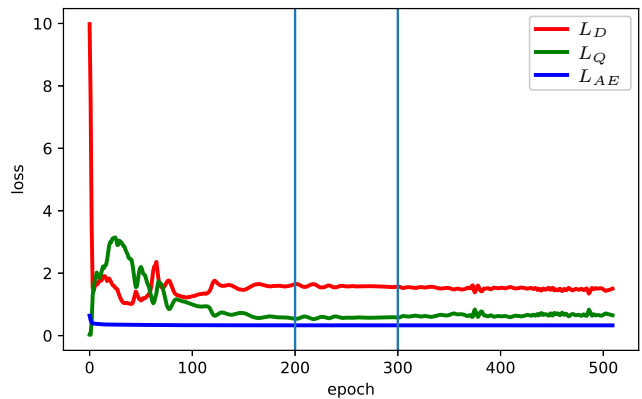}}
	\put(215,139){\scriptsize$L_D$}
	\put(215,129.7){\scriptsize$L_Q$}
	\put(215,121){\scriptsize$L_{AE}$}
\end{picture}
\caption{The change of AAE losses during first 500 training epochs. The training set includes normal walking gaits of 5 subjects. Our evaluation was performed on the epochs from 200 to 300.}
\label{fig:losses}
\end{figure}

As mentioned in~\cite{Goodfellow2014}, the GAN optimization attempts to converge to a saddle point instead of a minima, the loss is thus usually unstable during the training stage. Since there does not have an obvious criterion to stop training, we performed the evaluation on a range of 100 training epochs where the GAN-related losses were sufficiently stable. Concretely, we trained the AAE for 500 epochs and selected the models in a period of 100 epochs so that the losses did not suddenly change, an EER was then estimated for each AAE based on outputted measures, and the average EER was finally considered as an indicator of the method ability. A visualization of losses in our training stage is presented in Fig.~\ref{fig:losses}. The figure shows that the GAN losses were less stable after the $370^{\mathrm{th}}$ epoch, a range of 200-300 was thus selected. It is also obvious that the reconstruction loss $L_{AE}$ quickly converged  after a few epochs.

\subsection{Experimental results}\label{sec:results}
\begin{figure}[t]
\centering
\begin{picture}(250,278)
	\put(-2,0){\includegraphics[scale=0.52]{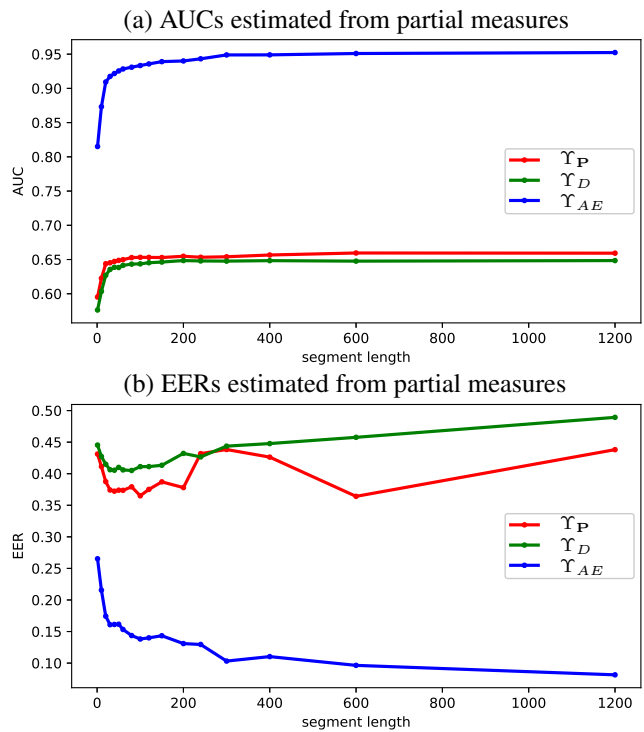}}
	\put(207,216.2){\scriptsize$\Upsilon_\mathbf{P}$}
	\put(207,208.1){\scriptsize$\Upsilon_D$}
	\put(207,200){\scriptsize$\Upsilon_{AE}$}
	\put(42,268){(a) AUCs estimated from partial measures}
	\put(207,78.0){\scriptsize$\Upsilon_\mathbf{P}$}
	\put(207,69.9){\scriptsize$\Upsilon_D$}
	\put(207,61.8){\scriptsize$\Upsilon_{AE}$}
	\put(42,130){(b) EERs estimated from partial measures}
\end{picture}
\caption{The average AUCs and EERs of the three partial measures estimated on segments of various lengths (including the per-frame assessment where the length is 1). The evaluation was performed according to the selected epoch period in Fig.~\ref{fig:losses}.}
\label{fig:resultsplit_raw}
\end{figure}

First, we consider the separation where the training and test sets respectively contain 5 and 4 subjects. Remember that our AAE was trained using only normal gaits. The ability of the three measures for the task of distinguishing normal and abnormal gaits is indicated in Fig.~\ref{fig:resultsplit_raw}. The reconstruction loss $\Upsilon_{AE}$ is a good measure since its EERs were low and quickly decreased when increasing the segment length. Therefore, $\Upsilon_{AE}$ should be used as the main factor in further combinations. The two others ($\Upsilon_\mathbf{P}$ and $\Upsilon_D$), however, are not individually good indicators since their EERs were very high and AUCs, low.

\begin{figure}[t]
\centering
\begin{picture}(290,345)
	\put(0,0){\includegraphics[width=0.65\textwidth]{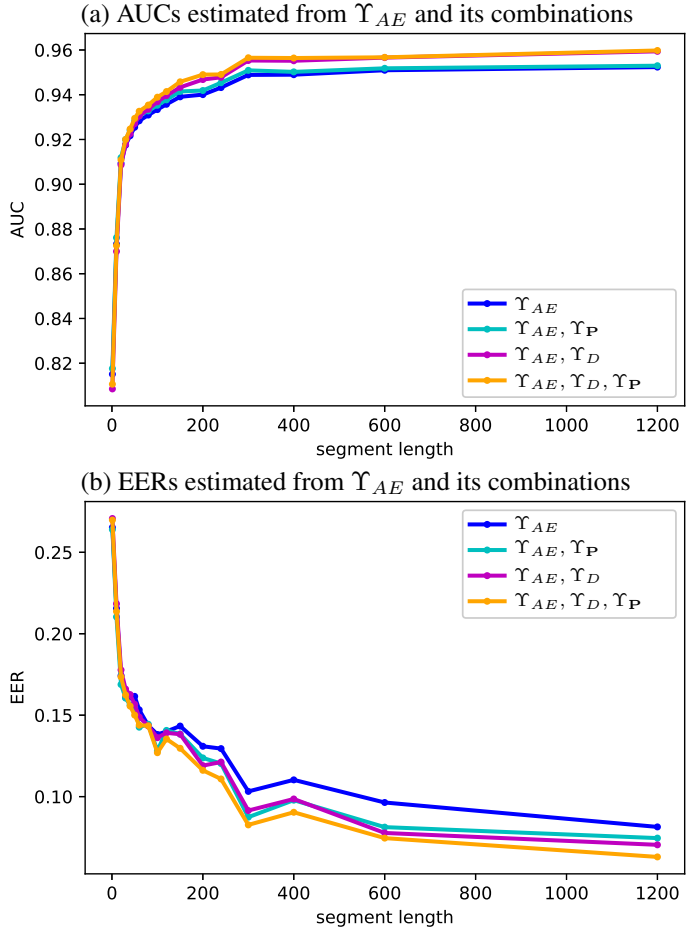}}
	\put(192,236){\scriptsize$\Upsilon_{AE}$}
	\put(192,227){\scriptsize$\Upsilon_{AE},\Upsilon_\mathbf{P}$}
	\put(192,217){\scriptsize$\Upsilon_{AE},\Upsilon_D$}
	\put(192,207){\scriptsize$\Upsilon_{AE},\Upsilon_D,\Upsilon_\mathbf{P}$}
	\put(28,345){(a) AUCs estimated from $\Upsilon_{AE}$ and its combinations}
	\put(192,153){\scriptsize$\Upsilon_{AE}$}
	\put(192,144){\scriptsize$\Upsilon_{AE},\Upsilon_\mathbf{P}$}
	\put(192,133.5){\scriptsize$\Upsilon_{AE},\Upsilon_D$}
	\put(192,124){\scriptsize$\Upsilon_{AE},\Upsilon_D,\Upsilon_\mathbf{P}$}
	\put(28,168){(b) EERs estimated from $\Upsilon_{AE}$ and its combinations}
\end{picture}
\caption{The average AUCs and EERs of $\Upsilon_{AE}$'s possible combinations estimated with different segment lengths. The AAE was evaluated according to the suggested 5:4 separation.}
\label{fig:resultsplit_comb}
\end{figure}

In order to enhance $\Upsilon_{AE}$ using the other measures, we attempted to perform some combinations. We observed that combining $\Upsilon_{AE}$ and the output of discriminator $\Upsilon_D$ decreased the EER while the opposite is true when we replaced $\Upsilon_D$ by $\Upsilon_\mathbf{P}$. We empirically found that this unwanted effect might be avoided when $\Upsilon_\mathbf{P}$ was raised by a small exponent (i.e. $\Upsilon_\mathbf{P} \leftarrow (\Upsilon_\mathbf{P})^u$ where $0<u<1$). The exponent only changes the contribution of $\Upsilon_\mathbf{P}$ in its combination, while its AUC and EER are still unchanged (see Fig.~\ref{fig:resultsplit_raw}) since the operation is monotonic. According to Fig.~\ref{fig:resultsplit_comb} (where $u = \frac{1}{8}$ after considering some small values), improving $\Upsilon_{AE}$ by both $\Upsilon_D$ and $\Upsilon_\mathbf{P}$ is recommended since its results were the best compared with the other combinations. Figure~\ref{fig:resultsplit_comb} also shows that the gait normality indicator tended to be better when using a higher value of temporal factor, i.e. estimating the gait index based on a longer sequence of point clouds.

\begin{figure}[t]
\centering
\begin{picture}(290,345)
	\put(0,0){\includegraphics[width=0.65\textwidth]{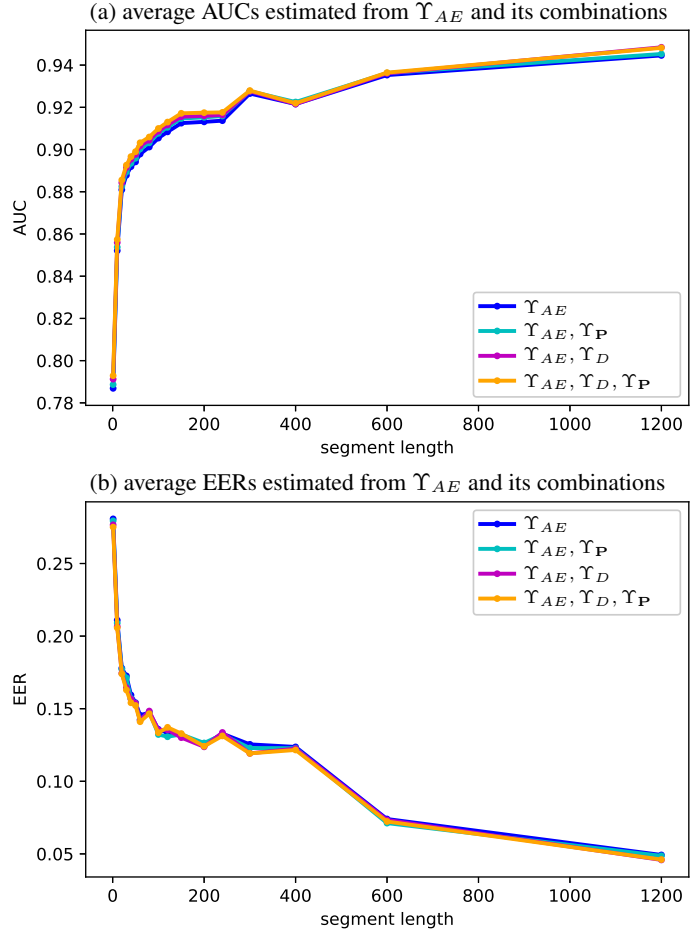}}
	\put(194,235){\scriptsize$\Upsilon_{AE}$}
	\put(194,226){\scriptsize$\Upsilon_{AE},\Upsilon_\mathbf{P}$}
	\put(194,217){\scriptsize$\Upsilon_{AE},\Upsilon_D$}
	\put(194,207){\scriptsize$\Upsilon_{AE},\Upsilon_D,\Upsilon_\mathbf{P}$}
	
	\put(30,346){\small (a) average AUCs estimated from $\Upsilon_{AE}$ and its combinations}
	\put(194,153){\scriptsize$\Upsilon_{AE}$}
	\put(194,144){\scriptsize$\Upsilon_{AE},\Upsilon_\mathbf{P}$}
	\put(194,134){\scriptsize$\Upsilon_{AE},\Upsilon_D$}
	\put(194,125){\scriptsize$\Upsilon_{AE},\Upsilon_D,\Upsilon_\mathbf{P}$}
	\put(30,168){\small (b) average EERs estimated from $\Upsilon_{AE}$ and its combinations}
\end{picture}
\caption{The average AUCs and EERs estimated in the leave-one-out evaluation stage. The discriminator output $\Upsilon_D$ slightly enhanced the reconstruction-based measure $\Upsilon_{AE}$.}
\label{fig:l1o}
\end{figure}

As for the leave-one-out (on subject) cross-validation, 9 AAEs were trained and evaluated according to 9 different data separations of ratio 8:1. AUCs and EERs are shown in Fig.~\ref{fig:l1o}. When combined with $\Upsilon_D$ and $\Upsilon_\mathbf{P}$, the reconstruction-based measure $\Upsilon_{AE}$ was slightly improved for assessing the gait normality. Let us notice that the selected epoch ranges of the 9 AAEs in the leave-one-out cross-validation were different depending on the stability of their training losses.

\subsection{Comparison}\label{sec:comparison}
\begin{table*}[t]
\centering
\footnotesize
\caption{Classification errors obtained from related studies and ours.}
\label{table:comparison}
\vspace{5pt}
\begin{tabular}{|c|g|g|g|g|g|}
\hline
\rowcolor{White}
 & & & \multicolumn{3}{l|}{Classification error ($\approx$ EER)} \\ \cline{4-6}
\rowcolor{White}
\multirow{-2}{*}{Data split} & \multirow{-2}{*}{Model} & \multirow{-2}{*}{Input type} & per-frame  & per-segment  & per-sequence  \\ \hline \hline 

 & HMM~\cite{Nguyen2016} & skeleton & - & 0.335 & 0.250 \\ \rowcolor{White}
 & One-class SVM~\cite{Bauckhage2009} & silhouette & 0.399 & 0.227 & 0.139 \\
 & HMM~\cite{Nguyen2018BHI} & depth map & - & 0.396 & 0.281 \\
 & Cross-correlation~\cite{Nguyen2018BHI} & silhouette & - & 0.381  & 0.250 \\
 & HMM + cross-correlation~\cite{Nguyen2018BHI} & depth map + silhouette & - & 0.377 & 0.218 \\ \rowcolor{White}
 & $\Upsilon_{AE}$ & point cloud & 0.265 & 0.153 & 0.081 \\ \rowcolor{White}
 & $\Upsilon_{AE}$ + $\Upsilon_\mathbf{P}$ & point cloud & \textbf{0.264} & \textbf{0.143} & 0.075 \\ \rowcolor{White}
 & $\Upsilon_{AE}$ + $\Upsilon_D$ & point cloud &  0.271 & 0.149 & 0.070 \\ \rowcolor{White}
\multirow{-9}{*}{\rotatebox[origin=c]{90}{5:4 separation}}  & $\Upsilon_{AE}$ + $\Upsilon_\mathbf{P}$ + $\Upsilon_D$ & point cloud & 0.270 & 0.144 & \textbf{0.063} \\ \hline \hline

 & HMM~\cite{Nguyen2016} & skeleton & - & 0.396 & 0.198 \\ \rowcolor{White}
 & One-class SVM~\cite{Bauckhage2009} & silhouette & 0.418 & 0.274 & 0.136 \\
 & HMM~\cite{Nguyen2018BHI} & depth map & - & 0.473 & 0.431 \\
 & Cross-correlation~\cite{Nguyen2018BHI} & silhouette & - & 0.321 & 0.097 \\
 & HMM + cross-correlation~\cite{Nguyen2018BHI} & depth map + silhouette & - & 0.319 & 0.083 \\ \rowcolor{White}
 & $\Upsilon_{AE}$ & point cloud & 0.281 & 0.145 & 0.049 \\ \rowcolor{White}
 & $\Upsilon_{AE}$ + $\Upsilon_\mathbf{P}$ & point cloud & 0.279 & 0.143 & 0.049 \\ \rowcolor{White}
 & $\Upsilon_{AE}$ + $\Upsilon_D$ & point cloud & 0.277 & 0.142 & 0.046 \\ \rowcolor{White}
\multirow{-9}{*}{\rotatebox[origin=c]{90}{leave-one-out}} & $\Upsilon_{AE}$ + $\Upsilon_\mathbf{P}$ + $\Upsilon_D$ & point cloud & \textbf{0.275} & \textbf{0.141} & \textbf{0.046} \\ \hline
\end{tabular}
\end{table*}

As mentioned in Section~\ref{sec:assessment}, related studies~\cite{Bauckhage2009,Nguyen2016,Nguyen2018BHI} were reimplemented and evaluated on our dataset under different input types.~\citet{Bauckhage2009} detected abnormal walking gaits based on a sequence of frontal silhouettes. The feature of each silhouette was extracted by fitting a lattice, and the posture was then described as a vector of some 2D corners that are pre-selected. The researchers embedded the temporal factor to improve their method by concatenating such consecutive vectors. The classification was performed using Support Vector Machines (SVMs) trained on multiple gait classes. Considering that objective under a different perspective, study~\cite{Nguyen2016} proposed another approach based on a sequence of 3D skeletons. The task of abnormal gait detection was performed according to an unsupervised (one-class) learning since defining specific abnormal gait types as in~\cite{Bauckhage2009} may reduce the generalization of the system in practical applications. Besides, the temporal factor was directly embedded in the stage of feature extraction. Concretely, the 3D skeleton in each frame was described by a vector of geometric quantities, and a sequence of such vectors corresponding to a gait cycle was then employed as a unit of gait representation. The gait index was provided by a HMM that described the change of postures within normal gait cycles. The method reported in~\cite{Nguyen2018BHI} estimated a gait normality index as a combination of two scores. The first one was determined by employing a HMM to measure the change of key points detected in consecutive depth maps. The second score was estimated by a cross-correlation on sequences of left and right projections of frontal silhouettes. The two scores were calculated with the support of a sliding window.

We reimplemented a HMM for~\cite{Nguyen2016}, a HMM and a cross-correlation procedure for~\cite{Nguyen2018BHI}. A one-class SVM was considered as a modification of the method~\cite{Bauckhage2009} to be used for a training set of only normal gait samples (similarly to~\cite{Nguyen2016,Nguyen2018BHI} and our work). The evaluation was also performed on the suggested separation in~\cite{NguyenReportDataset} as well as the leave-one-out cross-validation. Beside the assessment on a short sequence of frames (called \textit{per-segment}), i.e. feature concatenation of $\Delta=21$ consecutive frames for~\cite{Bauckhage2009}, automatically determined gait cycle for~\cite{Nguyen2016}, $\Delta=10$ frames within a sliding window for~\cite{Nguyen2018BHI} and $\Delta=60$ clouds for our method, we also considered the decision over the entire sequence of 1200 frames (so-called \textit{per-sequence}). The decision was determined by an alarm trigger in~\cite{Bauckhage2009}, smallest average log-likelihood of triple continuous cycles in~\cite{Nguyen2016}, and simply the mean score in~\cite{Nguyen2018BHI} as well as ours. Details of these results are shown in Table~\ref{table:comparison}.

The table shows that gait description over a long sequence was more reliable than considering short segments in all evaluated methods. The EERs resulting from $\Upsilon_{AE}$ and its combination with both $\Upsilon_\mathbf{P}$ and $\Upsilon_D$ were lower than the others in the leave-one-out cross-validation as well as in the per-sequence assessment according to the suggested separation. Let us notice the difference between the sequence-based assessments in~\cite{Bauckhage2009,Nguyen2016} and ours. Those two studies proposed non-linear computations on the per-segment results to obtain a reliable gait indicator. In other words, such segment-based measure might be noisy and the non-linear operations performed noise removal to keep a small piece of useful information. Unlike them, every per-frame measure in our work has an equal contribution to the index estimation. The method~\cite{Nguyen2018BHI} also used the same scheme but was affected by another drawback: the lack of generalization. This was clearly shown in Table~\ref{table:comparison} where its per-sequence EERs were significantly reduced in the leave-one-out evaluation compared with the case of testing on 4 subjects. The number of training subjects in the two cases was 8 and 5, respectively. Therefore, it is reasonable to guess that the method~\cite{Nguyen2018BHI} requires a large training dataset to provide a usable system. Recall that our AAE was designed with a simple architecture, we can thus expect to improve the model by carefully choosing component structures as well as tuning hyperparameters.

\begin{table*}[t]
\centering
\caption{Classification errors when evaluating gait index with the support of a sliding window.}
\label{table:overlapping}
\vspace{5pt}
\begin{tabular}{|l||l|l||l|l|}
\hline
\multirow{2}{*}{Model}  & \multicolumn{2}{c|}{5:4 separation} & \multicolumn{2}{c|}{leave-one-out} \\ \cline{2-5} 
 & $\Delta=10$ & $\Delta=21$ & $\Delta=10$ & $\Delta=21$ \\ \hline\hline
\rowcolor{Gray}
One-class SVM~\cite{Bauckhage2009} & - & 0.227 & - & 0.274 \\
HMM~\cite{Nguyen2018BHI} & 0.396 & - & 0.473 & - \\
Cross-correlation~\cite{Nguyen2018BHI} & 0.381 & - & 0.321 & - \\
HMM + cross-corr.~\cite{Nguyen2018BHI} & 0.377 & - & 0.319 & - \\\rowcolor{Gray}
$\Upsilon_{AE}$ & 0.211 & 0.174 & 0.207 & 0.169 \\\rowcolor{Gray}
$\Upsilon_{AE}$ + $\Upsilon_\mathbf{P}$ & \textbf{0.207} & \textbf{0.169} & 0.206 & 0.168 \\\rowcolor{Gray}
$\Upsilon_{AE}$ + $\Upsilon_D$ & 0.216 & 0.176 & 0.203 & 0.166 \\\rowcolor{Gray}
$\Upsilon_{AE}$ + $\Upsilon_\mathbf{P}$ + $\Upsilon_D$ & 0.213 & 0.171 & \textbf{0.202} & \textbf{0.165} \\ \hline
\end{tabular}
\end{table*}

Let us notice that the choice of segment length $\Delta=21$ and $\Delta=10$ respectively has a significant effect in~\cite{Bauckhage2009,Nguyen2018BHI} since these hyperparameters define the input of their models. Our approach, however, does not directly consider such temporal factor in the stage of model formation. Therefore, the per-segment evaluation of our method is an option where the segment length can be tuned depending on particular setup, objective, or application. These segments were non-overlapping to reduce the required computational cost. In order to emphasize the better ability of the proposed method compared with the others, a per-segment evaluation using sliding windows is presented in Table~\ref{table:overlapping}. This table shows that our method provided better results in describing gait index using a sliding window with small width. Notice that $\Delta=21$ and $\Delta=10$ were respectively recommended in~\cite{Bauckhage2009,Nguyen2018BHI} and were not optimal values for our approach. Therefore, a careful selection of such quantity is expected to improve our results (similarly to Fig.~\ref{fig:resultsplit_comb} and~\ref{fig:l1o}). Once again, the combination of the 3 measures provided best results in the phase of leave-one-out evaluation even with a very small window's width. 

\section{Conclusion}\label{sec:conclusion}
Adversarial auto-encoder and most GAN-based models have been employed for the task of data generation. This paper introduces another use of AAE to deal with a practical problem, i.e. gait index estimation in our work. The proposed approach focuses on the combination of measures provided from partial model components. The experiments demonstrate that an AAE has a great potential to work as a gait index estimator since such AAE with a very simple structure outperformed related studies that deal with various input types. The model can thus be expected to get better results when carefully tuning the architecture and related hyperparameters. Besides, finding a criterion for stopping the AAE training is also a significant work to extend our study. In addition, considering the underlying theory of combining different quantities ($\Upsilon_i$) could help to improve the ability of our system for the task of gait index estimation as well as for other similar applications.

\subsubsection*{Acknowledgments}

The authors would like to thank the NSERC (Natural Sciences and Engineering Research Council of Canada) for supporting this work (Discovery Grant RGPIN-2015-05671). We also thank Hoang Anh Nguyen (Airspace Systems Inc., CA, USA) for useful discussions.

\bibliography{references}
\bibliographystyle{iclr2019_conference}

\end{document}